\documentclass[letterpaper]{article} 
\usepackage{aaai25}  
\usepackage{times}  
\usepackage{helvet}  
\usepackage{courier}  
\usepackage[hyphens]{url}  
\usepackage{graphicx} 
\def\UrlFont{\rm}  
\usepackage{natbib}  
\usepackage{caption} 
\usepackage{algorithm}
\usepackage{algorithmic}

\usepackage{multirow}
\usepackage{booktabs}
\usepackage{amsmath,amssymb}
\usepackage{diagbox} 
\usepackage{mathrsfs}
\usepackage{makecell}
\usepackage{enumitem}
\usepackage[english]{babel} 
\usepackage{rotating}
\usepackage{soul, color, xcolor}
\usepackage{newfloat}
\usepackage{listings}
\DeclareCaptionStyle{ruled}{labelfont=normalfont,labelsep=colon,strut=off} 
\floatstyle{ruled}
\newfloat{listing}{tb}{lst}{}
\floatname{listing}{Listing}
\title{Low-Rank Adaptation with Task-Relevant Feature Enhancement for Fine-tuning Language Models}
\author {
    Changqun Li\textsuperscript{\rm 1},
    Chaofan Ding\textsuperscript{\rm 1},
    Kexin Luan\textsuperscript{\rm 1},
    Xinhan Di\textsuperscript{\rm 1},
}
\affiliations {
    \textsuperscript{\rm 1}AI Lab, Giant Network\\
    lichangqun@ztgame.com, dingchaofan@ztgame.com, luankexin@ztgame.com, dixinhan@ztgame.com
}

\usepackage{bibentry}

\begin{document}

\maketitle

\begin{abstract}
Fine-tuning pre-trained large language models in a parameter-efficient manner is widely studied for its effectiveness and efficiency. LoRA is one of the most widely used methods, which assumes that the optimization process is essentially low dimensional. Although LoRA has demonstrated commendable performance, there remains a significant performance gap between LoRA and full fine-tuning when learning new tasks. In this work, we propose Low-Rank Adaptation with Task-Relevant Feature Enhancement (LoRATRF) for enhancing task-relevant features from the perspective of editing neural network representations. To prioritize task-relevant features, a task-aware filter that selectively extracts valuable knowledge from hidden representations for the target or current task is designed. As the experiments on a vareity of datasets including NLU, commonsense reasoning and mathematical reasoning tasks demonstrates, our method reduces 33.71\% parameters and achieves better performance on a variety of datasets in comparison with SOTA low-rank methods.
\end{abstract}

\section{Introduction}
Pre-trained language models (PLMs) have shown remarkable performance across a wide variety of downstream natural language processing tasks through fine-tuning on task-specific labeled data~\citep{kenton2019bert,liu2019roberta,lewis2020bart}. However, fine-tuning all model parameters (\textit{full fine-tuning}) is prohibitively expensive. This issue is particularly salient with the ever-growing size of PLMs (e.g., BERT~\cite{kenton2019bert} with 330M parameters and GPT-3~\cite{brown2020language} with 175B parameters). 

To adapt general knowledge in pre-trained models to specific downstream tasks in a more parameter-efficient way, Parameter-Efficient Fine-Tuning (PEFT) methods have been proposed~\citep{houlsby2019parameter,pfeiffer-etal-2021-adapterfusion,li2021prefix,lester-etal-2021-power,ben-zaken-etal-2022-bitfit}. For example, \textit{adapter tuning}~\citep{houlsby2019parameter,pfeiffer-etal-2021-adapterfusion} inserts adapters to each layer of the pre-trained network. Inspired by the success of prompting methods that control PLMs through textual prompts~\cite{brown2020language}, \textit{prefix-tuning}~\cite{li2021prefix} and \textit{prompt-tuning}~\cite{lester-etal-2021-power} prepend an additional tunable prefix tokens to the input or hidden layers. Then, \textit{LoRA} and its derivatives~\citep{hu2022lora,zhang2023adaptive,liu2024dora} decomposes the attention weight gradients into low-rank matrices. Some studies~\citep{valipour-etal-2023-dylora,zhang2023adaptive,ding-etal-2023-sparse} mainly focused on dynamically adjusting the rank of LoRA in different layers. The above methods does not explore task-relevant features from the perspective of editing neural network representations.

In this paper, we take a step towards addressing the performance gap question, by proposing a new PEFT method called Low-Rank Adaptation with Task-Relevant Feature Enhancement (LoRATRF). We propose to enhance task-relevant features from the perspective of editing neural network representations. To prioritize task-relevant features, we introduce a task-aware filter that selectively extracts valuable knowledge from hidden representations for the target or current task. We conduct extensive experiments on a wide range of tasks and models to demonstrate the effectiveness of our method. To sum up, our contributions are:

\begin{itemize}
    \item We enhance the performance of LoRA from the perspective of neural network editing.
    \item We design a task-aware filter that can selectively extract valuable knowledge from the hidden representation of the current task, which can enhance the model's focus on crucial features.
\end{itemize}

\section{Related Work}
\subsection{Parameter-Efficient Fine-Tuning}
Parameter-efficient fine-tuning (PEFT) is an approach of optimizing a small number of parameters when fine-tuning a large pre-trained backbone model and keeping the backbone model untouched for adaptation~\cite{han2024parameter}. A branch of PEFT methods~\citep{lester-etal-2021-power,li-liang-2021-prefix,liu-etal-2022-p} is to add some special trainable vectors. Representative works in this direction are Prefix tuning~\cite{li-liang-2021-prefix}, Prompt tuning~\cite{lester-etal-2021-power} and P-tuning V2~\cite{liu-etal-2022-p}. Another approach~\citep{houlsby2019parameter,pfeiffer-etal-2021-adapterfusion,zhang-etal-2023-learned} is to insert additional neural modules to the backbone model, called Adapter. The reparameterization-based methods have attracted much attention~\cite{hu2022lora}. This type of PEFT method is closely related to intrinsic dimension~\citep{li2018measuring,aghajanyan-etal-2021-intrinsic}. However, the cost to train the above methods is large as the amount of trainable parameter is not small enough.

\subsection{LoRA and Its Variants}
LoRA~\cite{hu2022lora} is proven to be effective and yield stable results when applied to both relatively small pre-trained models and large language models~\citep{dettmers2023qlora,hu-etal-2023-llm}. Then, some researchers explore more flexible and appropriate ranks, such as DyLoRA, AdaLoRA, and SoRA. Other works focus on the combination of LoRA and other approaches, such as AdaMix~\cite{wang-etal-2022-adamix} and QLoRA~\cite{dettmers2023qlora}. Besides, LoRAHub~\cite{huang2023lorahub} and LoRAMoE~\cite{wu2023mole} focus on how to merge multiple LoRA blocks that are fine-tuned on different tasks respectively. However, the above methods lacks efficiency on low-rank representation of complex reasoning tasks for LLMs.



\begin{figure}[htp]
\centering
\includegraphics[scale=0.55]{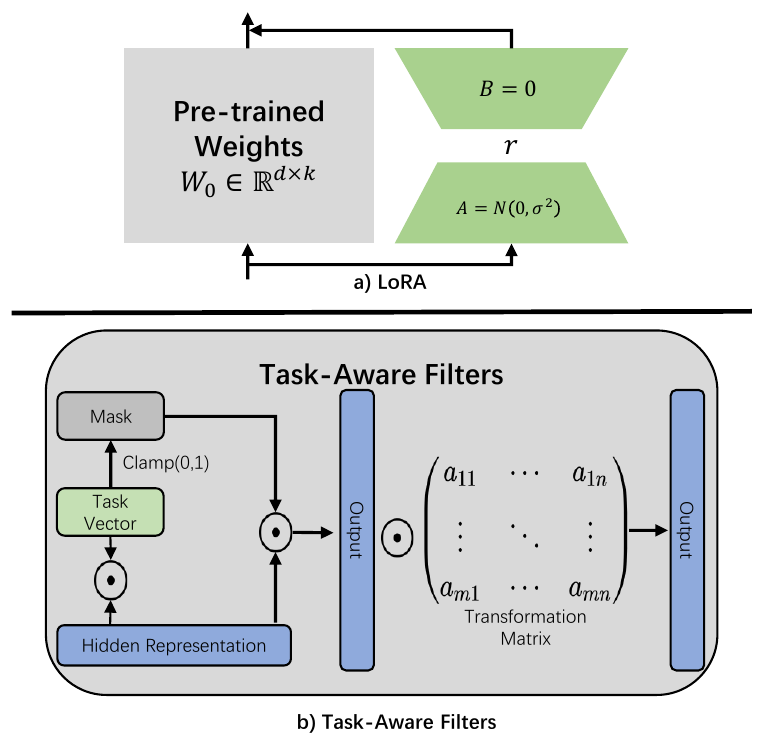}
\caption{\label{model}
The overview of our approach. Task-aware filter can selectively extract valuable knowledge from hidden representations for the target task. $\odot$ refers to element-wise multiplication.}
\end{figure}

\section{Methodology}
\subsection{Preliminaries}
\paragraph{Low-Rank Adaptation (LoRA).} LoRA~\cite{hu2022lora} approximates the incremental update by decomposing it into the product of two low-rank matrices, constraining the update to a low-rank space and making the fine-tuning process more efficient. Through this approximate low-rank decomposition, we have:
\begin{equation}
    \Delta W \approx BA,
\end{equation}
where $B \in R^{m \times r}$, $A \in R^{r \times n}$, and the rank $r << min(m,n)$. In the forward propagation of LoRA, for an input representation $x$, the output after passing through the parameter matrix $W_{0}$ is
\begin{equation}
    h=W_{0}x + \Delta W x = W_{0}x + BAx
\end{equation}
To ensure that introducing LoRA at the initial stage does not impact the computation results of the model’s forward propagation, it is crucial to ensure that $BAx$ = 0. To achieve this, LoRA initializes $A$ as a random Gaussian matrix and $B$ as a zero matrix. During training, $W_{0}$ is frozen, and $B$ and $A$ are treated as trainable parameters. Upon completion of training, the parameter matrices $A$ and $B$ are merged into $W_{0}$ to form the final parameter matrix
\begin{equation}
    W_{ft} = W_{0} + BA
\end{equation}
It is noteworthy that the final update is $BA$, which is constrained within a low-rank space.


\subsection{Motivation}


To further improve performance, we propose Low-Rank Adaptation with Task-Relevant Featur Enhancement (LoRATRF). Figure~\ref{model} gives an overview of our approach, where the task-aware filter identifies task-relevant features within hidden representations and adaptively integrates these features back into the representations.



\begin{table*}[t]
\centering
\begin{tabular}{l|ccccccccc|c}
\Xhline{1.2pt} 
\multirow{2}*{\textbf{Method}} & \multirow{2}*{\textbf{\# Params}} & \textbf{MNLI} &  \textbf{SST-2} &  \textbf{CoLA} & \textbf{QQP} & \textbf{QNLI} & \textbf{RTE} & \textbf{MRPC} & \textbf{STS-B} & \textbf{All} \\ 
& & Acc & Acc & Mcc & Acc/F1 & Acc & Acc & Acc & Corr & Ave. \\ \hline
Full FT &  {184M} & {89.90} & {95.63} & {69.19} & {\bf92.40/89.80} & {94.03} & {83.75} & {89.46} & {91.60} & {88.42}  \\ \hline
BitFit & {0.1M} & {89.37} & {94.84} & {66.96} & {88.41/84.95} & {92.24} & {78.70} & {87.75} & {91.35} & {86.06}  \\ \hline
HAdapter & {1.22M} & {90.13} & {95.53} & {68.64} & {91.91/89.27} &  {94.11} & {84.48} & {89.95} & {91.48} & {88.39} \\ 
PAdapter & {1.18M} & {90.33} & {95.61} & {68.77} & {92.04/89.40} & {94.29} & {85.20} & {89.46} & {91.54} & {88.52} \\ 
LoRA$_{r=8}$ & {1.33M} & {90.65} & {94.95} & {69.82} & {91.99/89.38} & {93.87} & {85.20} & {89.95} & {91.60} & {88.60} \\
AdaLoRA & {1.27M} & {90.76} & {\bf 96.10} & {71.45} & {92.23/89.74} & {94.55} & {\bf 88.09} & {90.69} & {91.84} & {89.49} \\
LoRATRF & 0.88M & \textbf{90.87} & 95.99 & \textbf{71.61} & 92.00/89.75 & \textbf{94.69} & 87.73 & \textbf{91.18} & \textbf{91.89} & \textbf{89.52}  \\ \Xhline{1.2pt} 
\end{tabular}
\caption{ Results with DeBERTaV3-base on GLUE development set. Best performances are highlighted in bold. Full FT, HAdapter and PAdapter represent full fine-tuning, Houlsby adapter, and Pfeiffer adapter respectively. We report baseline results directly from~\cite{zhang2023adaptive}. We run the experiment on 5 different random seeds and report the mean.}
\label{glue_result}
\end{table*}

\subsection{Task-Aware Filters}
\label{task_aware_section}
Firstly, we introduce task-aware filters ~\cite{zou2023representation} that are able to select task-relevant features in the output and then reincorporate them into the output representation.

Specifically, we design a learnable task vector $t_{\xi} \in R^{d}$ and employ it to perform the matrix product with the representation vector $h^{l}$ in each Transformer layer $l$, subsequently, the resulting product is clamped to [0,1]. This selective mechanism preferentially retains tokens that exhibit high similarity to $t_{\xi}$, while effectively attenuating others via a soft masking procedure.

\begin{equation}
    \tilde h_{i}^{l} = sim(h_{i}^{l}, t_{\xi}) \cdot h_{i}^{l}
\end{equation}
Here $t_{\xi}$ acts as a task embedding that encodes what kind of tokens are important for the task, and each token $h_{i}^{l}$ is reweighted by its relevance (measured by cosine similarity) with the task embedding, thereby simulating token prioritization based on task-related importance.

Secondly, we integrate a transformation matrix $T \in R^{d \times d}$ to execute linear transformations on the reweighted representation~\footnote{To promote parameter efficiency, we approximate the transformation matrix $T$ using the product of two low-rank matrices.}, permitting finer adjustments and augmentations that are tailored to the task:

\begin{equation}
    \hat h_{i}^{l} = \tilde h_{i}^{l} \cdot T
\end{equation}
The transformation matrix $T$ is designed to linearly transform the selected and reweighted token representations in a task-adaptive manner. This approach ensures that the resulting representation is discriminative and adaptable, prioritizing the information elements that is most relevant to the task. Finally, this refined information $\hat h_{i}^{l}$ is added to the original representation to achieve comprehensive enhancement.



\section{Experimental Setup}
\subsection{Datasets}
\paragraph{General Language Understanding Evaluation:}GLUE~\cite{wang2019glue} is a generalized natural language understanding assessment benchmark that includes a variety of tasks such as natural language inference, sentiment analysis, and sentence similarity evaluation, from which we select eight tasks for systematic evaluation, including Corpus of Linguistic Acceptability (CoLA), Multi-Genre Natural Language Inference (MNLI), Microsoft Research Paraphrase Corpus (MRPC), Question Natural Language Inference (QNLI), Quora Question Pairs (QQP), Recognizing Textual Entailment (RTE), Stanford Sentiment Treebank (SST-2), Semantic Textual Similarity Benchmark (STS-B).

\paragraph{Mathematical Reasoning:} (1) \textbf{GSM8K}~\cite{cobbe2021training} dataset consists of high quality linguistically diverse grade school math word problems, (2) \textbf{SVAMP}~\cite{patel-etal-2021-nlp} benchmark consists of one-unknown arithmetic word problems, (3) \textbf{AddSub}~\cite{hosseini-etal-2014-learning} is a specialized dataset designed for evaluating algorithms.

\paragraph{Commonsense Reasoning:} (1) BoolQ~\cite{clark-etal-2019-boolq} dataset is a question-answering dataset for yes/no questions containing 15942 examples. (2) PIQA~\cite{bisk2020piqa} dataset of questions with two solutions requiring physical commonsense to answer; (3) SIQA~\cite{sap-etal-2019-social} focuses on reasoning about people’s actions and their social implications; (4) HellaSwag~\cite{zellers-etal-2019-hellaswag} is a challenging dataset, which contains questions to select the best endings to complete sentences.

\begin{table*}[htp]
\centering
\begin{tabular}{ccccccccc}
\toprule
\textbf{LLM} & \textbf{Method} & \textbf{GSM8K} & \textbf{AddSub} & \textbf{SVAMP} & \textbf{HellaSwag} & \textbf{BoolQ} & \textbf{PIQA} & \textbf{SIQA} \\ \hline
\textbf{GPT-3.5} & & 56.4 & 85.3 & 69.9 & 78.5 & 73.1 & 85.4 & 68.5 \\ \hline
\multirow{6}{*}{\textbf{LLaMA-7B}} & Prefix & 24.4 & 57.0 & 38.1 & 42.1 & 64.3 & 76.8 & 73.9 \\
& Series & 33.3 & 80.0 & 52.3 & 67.9 & 63.0 & 79.2 & 76.3 \\
& Parallel & 35.3 & \textbf{86.6} & 49.6 & 69.8 & 67.9 & 76.4 & 78.8 \\
& LoRA & 37.5 & 83.3 & 52.1 & 78.1 & 68.9 & 80.7 & 77.4 \\
& DoRA & 38.4 & 84.2 & 52.7 & \textbf{84.8} & 68.5 & 82.9 & \textbf{79.6} \\
& LoRATRF & \textbf{38.6} & 84.1 & \textbf{53.0} & 82.4 &  \textbf{69.6} & \textbf{83.5} & 78.4\\
\bottomrule
\end{tabular}
\caption{Comparison results of different methods based on LLaMA-7B on reasoning datasets. We report some baseline results directly from~\cite{liu2024dora} and~\cite{hu-etal-2023-llm}.}
\label{llama}
\end{table*}

\subsection{Baselines} 
We compare our methods to Full fine-tuning, Bitfit, Adapter tuning, LoRA and AdaLoRA. \textbf{Bitfit}~\cite{ben-zaken-etal-2022-bitfit} fine-tunes bias vectors. \textbf{Houlsby adapter} ~\cite{houlsby2019parameter} is inserted between the self-attention module and the FFN module. \textbf{Pfeiffer adapter}~\cite{pfeiffer-etal-2021-adapterfusion} inserts the adapter after FFN modules and LayerNorm modules. \textbf{LoRA}~\cite{hu2022lora} parameterizes incremental updates by two small matrices. \textbf{AdaLoRA}~\cite{zhang2023adaptive} expresses the low-rank multiplication of LoRA. In empirical, we find that applying LoRA to $W{v}$, $W_{f_{1}}$ and $W_{f_{2}}$ matrices can achieve the best performance (Please see \textbf{Different Choices of Modules to Adapt
} Section).



\subsection{Implementation Details}
We implement our method for fine-tuning DeBERTaV3-base~\cite{he2022debertav3} and large language model LLaMA-7B~\cite{touvron2023llama}. LoRA~\cite{hu2022lora} scales $\Delta W$ by $\alpha/r$ where $\alpha$ is a constant in $r$. As a result, the magnitude of output can be consistent given different $r$. It reduces the efforts of retuning learning rate when varying $r$. Typically $\alpha$ is set as 16 or 32 and never tuned. 

\subsection{Evaluation}
For the GLUE benchmark, we report both accuracy and F1 for QQP in GLUE. For STS-B, we report the average correlation. For CoLA, we report Matthews correlation. For all remaining sub-tasks in GLUE, we report accuracy. For mathematical and commonsense reasoning datasets, we report accuracy.

\section{Main Results}
\subsection{Natural Language Understanding}
We compare LoRATRF with various baselines. Table~\ref{glue_result} shows experimental results on the GLUE development set. We see that LoRATRF achieves better or on par performance compared with existing approaches on all datasets. For example, our method attains an accuracy of 71.61\% on CoLA, surpassing AdaLoRA baseline by $0.2\%$, all while utilizing fewer parameters (0.88M compared to 1.27M). Among the baseline methods, the AdaLoRA method performs the best, which may be because it designs a method to dynamically allocate the rank of LoRA in different layers based on their importance. For all tasks except QQP, our method demonstrates different degrees of performance enhancement. These experiments verify the general applicability of our method to the NLU tasks.

\subsection{Commonsense and Mathematical Reasoning}
We also conduct experiments on the Mathematical and Commonsense Reasoning task using LLaMA-7B~\cite{touvron2023llama}. We use the same rank size (r=32) as the baseline method and only apply the LoRA module to output projection ($W_{o}$) in the self-attention, and two weight matrices ($W_{f_{1}}, W_{f_{2}}$) in two-layer FFNs. Comparison results are reported in Table~\ref{llama}. Notably, in the LLaMA-7B model, where DoRA exceeds the performance of other baselines, which may be due to DoRA decomposes weights for enhanced learning capacity. LoRATRF attains an accuracy of 85.6\% on HellaSwag, surpassing DoRA baseline by 0.8\%. To sum up, LoRATRF achieves the best performance in GSM8K. The results show that our method maintains its effectiveness in LLM, and further illuminate the significance of enhancing task-related features.



\section{Quantitative Analysis}

\subsection{Different Choices of Modules to Adapt}
\label{different_adapted_modules_section}
\begin{figure}[htp]
\centering
\includegraphics[scale=0.31]{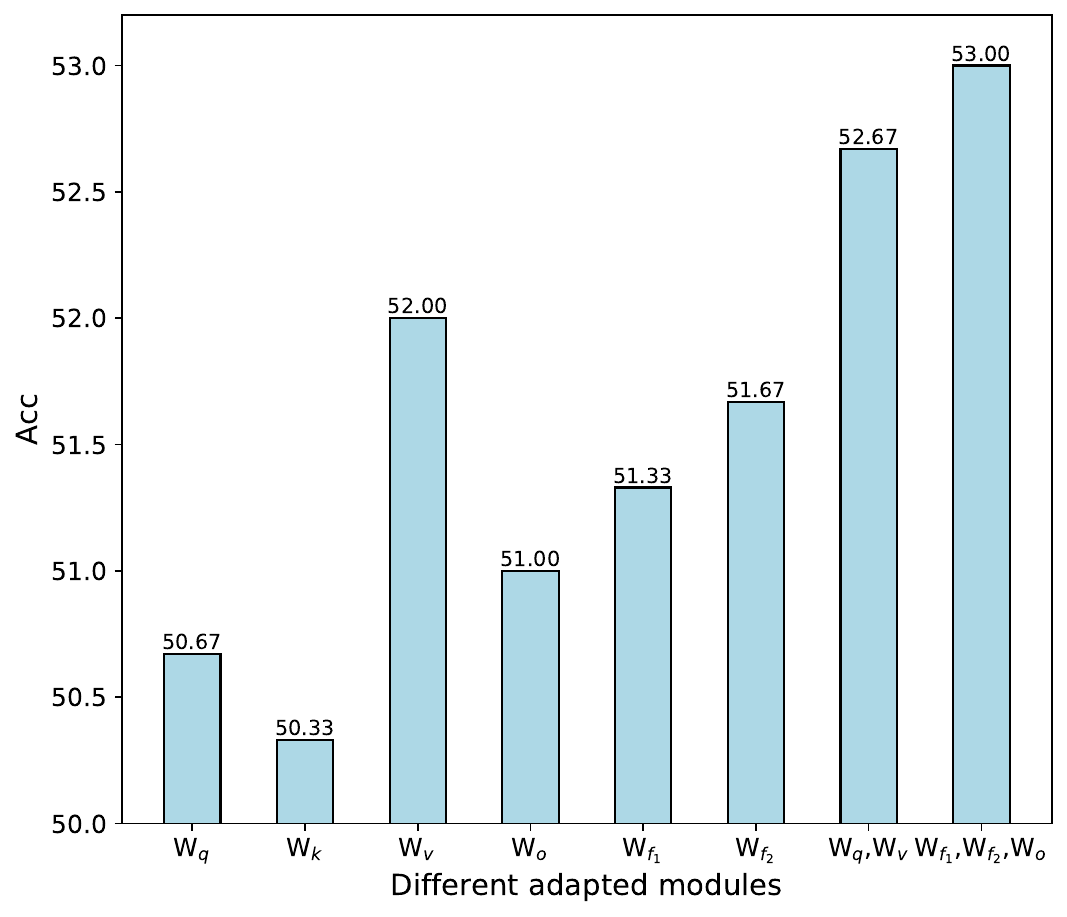}
\caption{\label{different_adapted_modules}
Testing performance of LLaMA-7B on SVAMP with different adapted modules.}
\end{figure}

We study the choices of modules to adapt for our method on SVAMP. We choose possible modules to adapt within query/key/value projection ($W_{q}, W_{k}, W_{v}$), output projection ($W_{o}$) in the self-attention, and two weight matrices ($W_{f_{1}}, W_{f_{2}}$) in two-layer FFNs. We hold the number of trainable parameters at the same level. Figure~\ref{different_adapted_modules} shows the performance when fine-tuning specific modules, which demonstrates that adapting $W_{f_{1}}$, $W_{f_{2}}$ and $W_{o}$ yields the highest performance. 

\subsection{Robustness of LoRATRF towards different rank settings}
\begin{figure}[htp]
\centering
\includegraphics[scale=0.31]{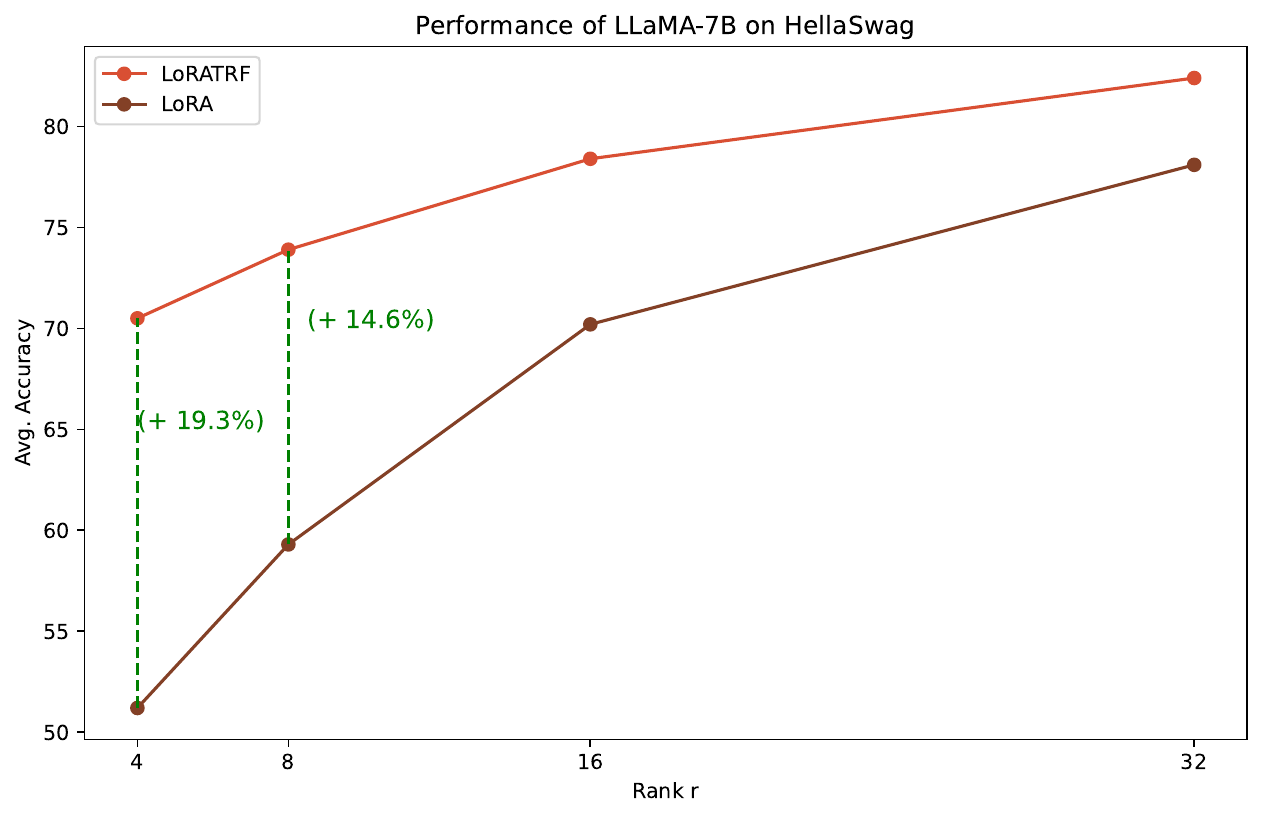}
\caption{\label{different_rank_settings}
Testing performance of LLaMA-7B on HellaSwag with different rank settings.}
\end{figure}

This section explores the impact of various rank configurations on LoRATRF and LoRA by adjusting $r$ within the set \{4, 8, 16, 32\} and assessing the performance of the fine-tuned LLaMA-7B on commonsense reasoning dataset HellaSwag. The average accuracies of LoRA and LoRATRF across different ranks are depicted in Figure~\ref{different_rank_settings}. From Figure~\ref{different_rank_settings}, we can observe that LoRATRF consistently surpasses LoRA across all rank configurations. Notably, the performance gap widens for ranks below 8, where LoRA’s average accuracies drop to 59.3\% for $r$ = 8 and 51.2\% for $r$ = 4. In contrast, LoRATRF retains a notable accuracy of 73.9\% for $r$ = 8 and 70.5\% for $r$ = 4, demonstrating its resilience and consistently superior performance over LoRA regardless of the rank setting.

\section{Conclusion}
In this work, we propose Low-Rank Adaptation with Task-Relevant Feature Enhancement, a novel approach aimed at bridging the performance gap between LoRA and full fine-tuning on complex tasks. We introduce task-aware filters to improve performance by prioritizing task-relevant features. Experiments on diverse benchmarks with different settings confirm the effectiveness of our method.

\clearpage
\bibliography{aaai25}

\end{document}